\documentclass[letterpaper, 10 pt, conference]{ieeeconf}  

\IEEEoverridecommandlockouts                              

\renewcommand{\baselinestretch}{0.99}

\usepackage{graphicx}
\usepackage{booktabs}
\usepackage{tabularx}
\usepackage{todonotes}

\usepackage{multicol}
\usepackage[bookmarks=true]{hyperref}
\usepackage{multirow}
\usepackage{times}
\usepackage{epsfig}
\usepackage{graphicx}
\usepackage{amsmath}
\usepackage{amssymb}
\makeatletter
\let\NAT@parse\undefined
\makeatother
\usepackage[numbers]{natbib}

\usepackage{booktabs}
\usepackage{tabularx}
\usepackage[font=small,labelfont=bf]{caption}

\usepackage[font=small]{caption}

\usepackage{hyperref}
\usepackage[capitalise, nameinlink]{cleveref}

\newcommand{\fig}[1]{Fig.~\ref{fig:#1}}

\newcommand{\sect}[1]{Sec.~\ref{sec:#1}}
\newcommand{\algname}{RoboVQA}
\newcommand{\modelname}{\algname-VideoCoCa}
\newcommand{\website}{\href{https://robovqa.github.io}{robovqa.github.io}}

\title{\LARGE \bf
\algname{}: Multimodal Long-Horizon Reasoning\\ for Robotics
}

\author{
Pierre Sermanet,
Tianli Ding,
Jeffrey Zhao,
Fei Xia,
Debidatta Dwibedi,
Keerthana Gopalakrishnan,\\
Christine Chan,
Gabriel Dulac-Arnold,
Sharath Maddineni,
Nikhil J Joshi,
Pete Florence,
Wei Han,\\
Robert Baruch,
Yao Lu,
Suvir Mirchandani, 
Peng Xu,
Pannag Sanketi,
Karol Hausman,\\
Izhak Shafran,
Brian Ichter,
Yuan Cao\\
\texttt{Google DeepMind}
}

\usepackage{fancyhdr}
\begin{document}
\maketitle

\begin{abstract}
    We present a scalable, bottom-up and intrinsically diverse data collection scheme that can be used for high-level reasoning with long and medium horizons and that has 2.2x higher throughput compared to traditional narrow top-down step-by-step collection. We collect realistic data by performing any user requests within the entirety of 3 office buildings and using multiple embodiments (robot, human, human with grasping tool). With this data, we show that models trained on all embodiments perform better than ones trained on the robot data only, even when evaluated solely on robot episodes. We explore the economics of collection costs and find that for a fixed budget it is beneficial to take advantage of the cheaper  human collection along with robot collection. We release a large and highly diverse (29,520 unique instructions) dataset dubbed \algname{} containing 829,502 (video, text) pairs for robotics-focused visual question answering. We also demonstrate how evaluating real robot experiments with an intervention mechanism enables performing tasks to completion, making it deployable with human oversight even if imperfect while also providing a single performance metric.
    We demonstrate a single video-conditioned model named \modelname{} trained on our dataset that is capable of performing a variety of grounded high-level reasoning tasks in broad realistic settings with a cognitive intervention rate 46\% lower than the zero-shot state of the art visual language model (VLM) baseline and is able to guide real robots through long-horizon tasks. The performance gap with zero-shot state-of-the-art models indicates that a lot of grounded data remains to be collected for real-world deployment, emphasizing the critical need for scalable data collection approaches. Finally, we show that video VLMs significantly outperform single-image VLMs with an average error rate reduction of 19\% across all VQA tasks. Thanks to video conditioning and dataset diversity, the model can be used as general video value functions (e.g. success and affordance) in situations where actions needs to be recognized rather than states, expanding capabilities and environment understanding for robots.
    Data and videos are available at \website
\end{abstract}



\begin{figure*}[h]
  \centering
  \includegraphics[width=1\linewidth]{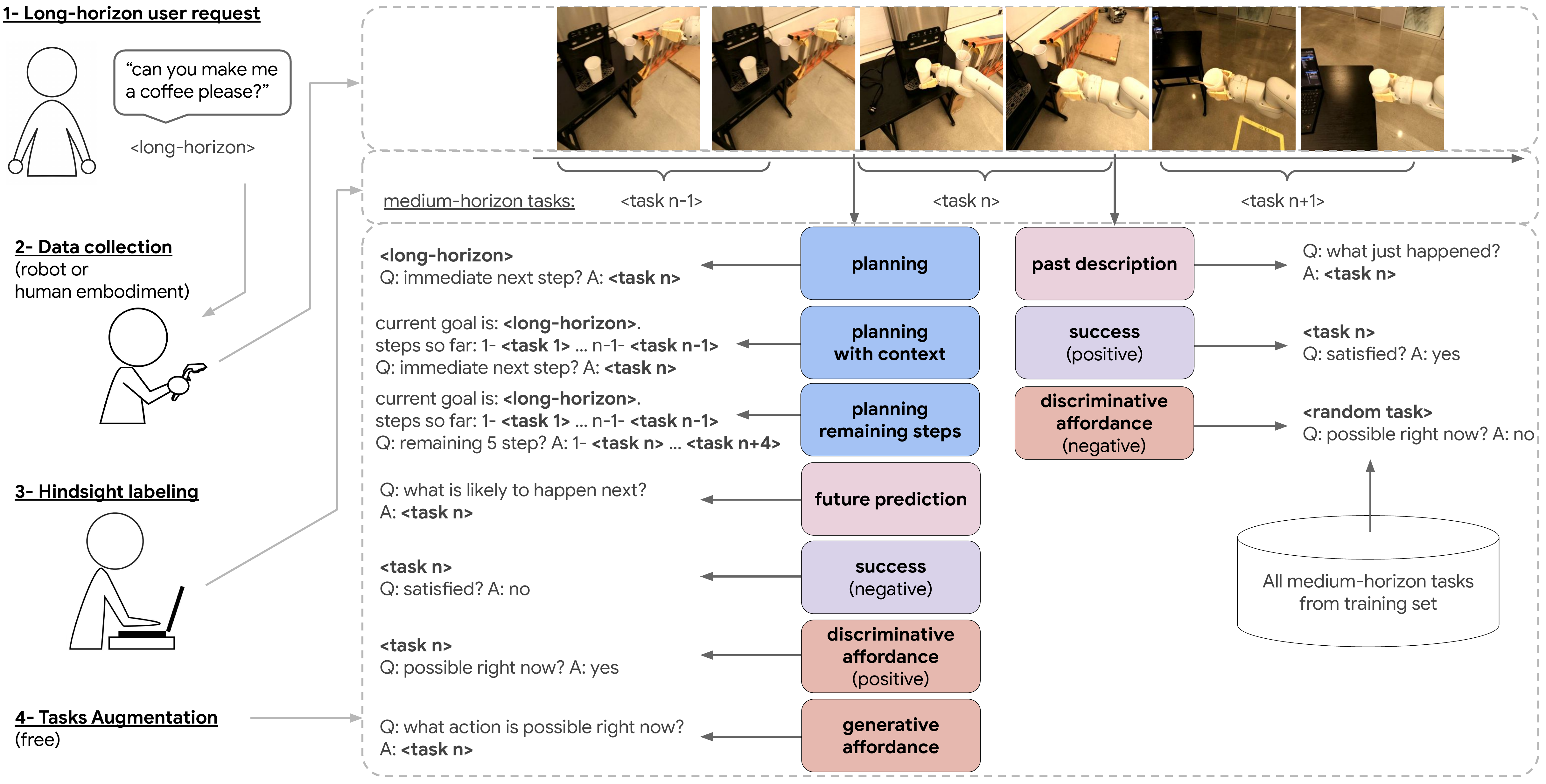}  
\caption{\textbf{Data collection procedure}: Given long-horizon user requests, a human operator teleoperates a robot to fulfill the task. Medium-horizon tasks are then labeled in hindsight via crowd-sourcing, with temporal segmentation and task instruction for each segment. Finally, from a sequence of labeled segments, we automatically generate 10 types of question/answer pairs.}
  \label{fig:data_collection}
\end{figure*}

\section{Introduction}
\label{sec:intro}

The field of textual high-level reasoning has seen major breakthroughs recently with large language models (LLMs)~\cite{OpenAI2023GPT4TR,chowdhery2022palm}, while progress has also been made in visual language models (VLMs)~\cite{driess2023palme}, high-level reasoning that is grounded in the real world remains a challenging task and critical for robotics. Can the state-of-the-art VLMs trained on available multimodal datasets perform grounded tasks with high accuracy in the real-world? We aim to answer the question by showing that new large scale data collection are still needed to achieve lower error rates outside of lab environments.
A major difficulty for VLMs stems from the high-dimensionality of the real world which, accordingly requiring large amounts of multimodal data (video, language, actions) for training. Hence a major contribution of our work is to validate more efficient data collection approaches than the traditional top-down step-by-step collection~\cite{rt12022arxiv}, by reducing overheads such as resets and scene preparations and leveraging the low costs of human embodiment collection. With a crowd-sourced bottom-up approach where long-horizon tasks are decided by real users the resulting medium-horizon steps are naturally highly diverse, relevant and on-distribution for users. Not only it is a more efficient way to collect medium-horizon steps, we also get long-horizon coherent sequences which can train models to perform planning tasks. With a 2.2x throughput increase compared to the traditional method, it is preferable to collect data this way even if long-horizon tasks are not needed.
While we do collect robot actions in this dataset, the focus of this paper is on high-level reasoning tasks, we can hence train on embodiments which do not come with motor commands and observe transfer of knowledge between embodiments. We find in \sect{expemb} that for a fixed collection budget, it is beneficial for high-level reasoning to jointly with cheaper human embodiment even when evaluating on the robot embodiment only.

Our contributions can be summarized as follows:
\begin{enumerate}
    \item We demonstrate a scalable, bottom-up and intrinsically diverse data collection scheme that can be used for high-level reasoning with long and medium horizons and that has 2.2x higher throughput compared to traditional narrow top-down step-by-step collection and show additional cheap human embodiment data improves performance.
    \item We release a large and diverse cross-embodiment dataset of 829,502 (video, text) pairs for robotics-focused visual question answering.
    \item We demonstrate a single video-conditioned model trained on the dataset that is capable of performing a variety of tasks with higher accuracy than baselines and is able to guide real robots through long-horizon tasks.
    \item We establish a robotics VQA benchmark and long-horizon planning benchmark with an intervention mechanism on real robots providing a single performance metric and enabling performing tasks to completion, making it deployable with human oversight even when imperfect.
\end{enumerate}

\section{Data} \label{sec:data}

\begin{figure*}[h!]
  \centering
  \includegraphics[width=1\linewidth]{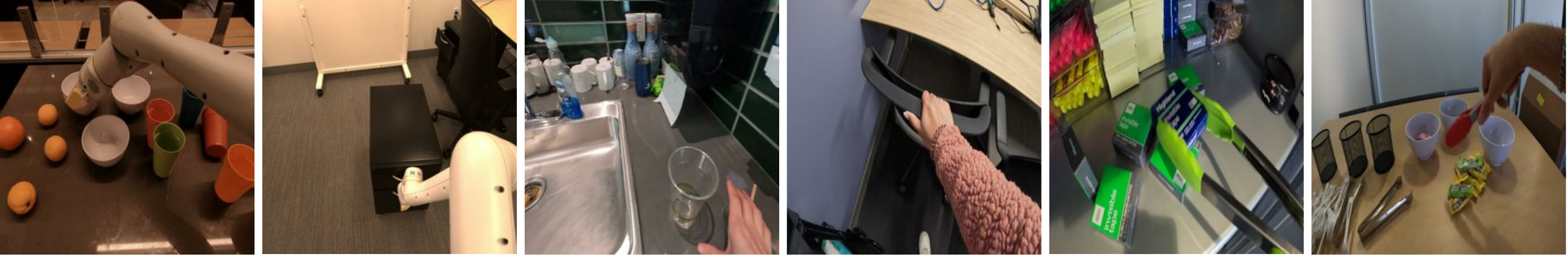}  
  \caption{\small{Examples of 3 embodiments in the dataset: robot, human (single) arm, human using a grasping tool.}}
  \label{fig:embodiments}
\end{figure*}

\textbf{Collection \& Dataset:} In \fig{data_collection} we describe the collection process, from user request to VQA tasks generation. We collect episodes from any long-horizon tasks within the entirety of 3 office buildings and with 3 embodiments (\fig{embodiments}), resulting in 238 hours of video (10 days), 5,246 long-horizon episodes and 92,948 medium-horizon episodes. The average long-horizon episode lasts 102 seconds, the medium-horizon average is 14s. Because evaluation of freeform text answers are performed by humans in our experiments, we keep the validation and test sets small on purpose with approximately 1,000 VQA entries for each (coming from 50 episodes each). While there can be overlap in scenes between training and val/test, there is no overlap in episodes. For more statistics, see \sect{dataset_statistics}.

\textbf{Task diversity:} To ensure that our dataset and benchmark do not overfit to a specific environment, domain or task, we collect examples over a wide range of tasks compared to more traditional collections~\citep{saycan2022arxiv} where a fixed and small list of tasks is decided in advance by researchers and engineers in a top-down fashion. We opt for a bottom-up approach where a large number of tasks are crowd-sourced by users and tele-operators. This favors breadth and a better alignment with a distribution of requests coming from real users. This results in high tasks diversity (26,798 unique medium-horizon instructions, 2,722 unique long-horizon instructions).

\begin{figure}[htb!]
  \centering
  \includegraphics[width=1\linewidth]{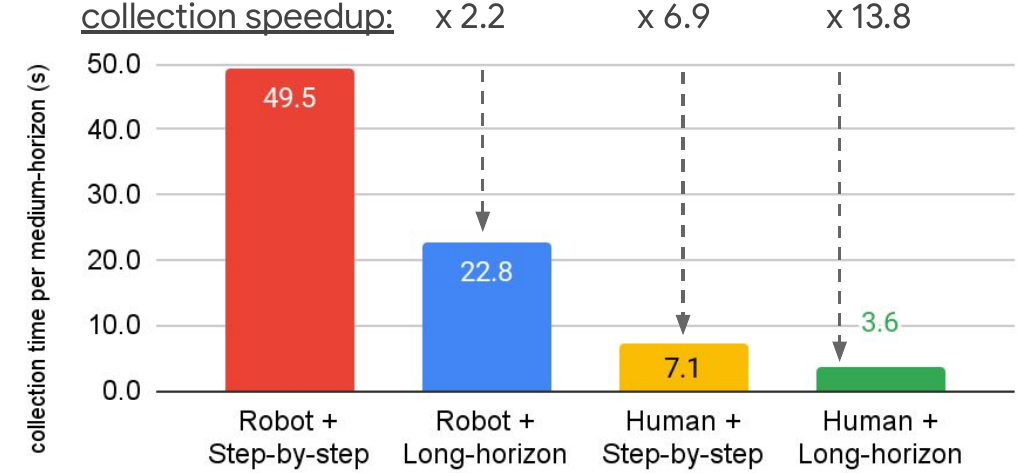}  
\caption{Throughput gains compared to the traditional top-down step-by-step collection approach. The throughput of our long-horizon collection is 2.2x higher for robot collection and 13.8x higher with human bodies (compared to the robot used in our experiments).}
  \label{fig:throughput}
\end{figure}

\textbf{Throughput and costs:} Much of the throughput gains reported in \fig{throughput} come from collecting medium-horizon episodes in a continuous fashion without needing to reset the scene or the robot. Note that the hindsight labeling process can be parallelized via crowd-sourcing and does not impact the throughput if performed in parallel, however it remains a cost in the collection budget. The VQA tasks however are generated for free by taking advantage of the known sequence of past and future tasks and positioning the questions in time with respect to different known semantic points (e.g. before or after a medium-horizon task was performed).

\begin{figure*}[h!]
  \centering
  \includegraphics[width=1\linewidth]{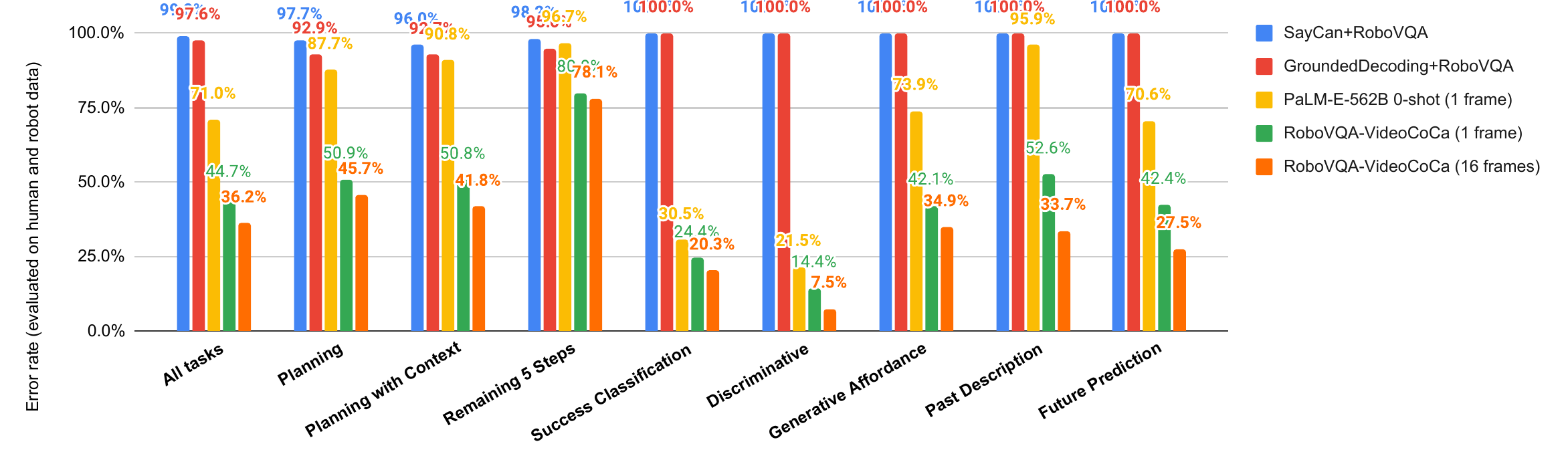}
  \caption{\small{\textbf{VQA Error rates}: we evaluate all models on the test set using human raters. We observe that state-of-the-art methods do not perform well in realistic settings in zero-shot, thus motivating the need for further scalable data collections. We also observe substantial gains when using video (16 frames) vs image conditioning.
  }}
  \label{fig:vqa_eval_all}
\end{figure*}

\textbf{Chain-of-Thought:} Decomposing high-level goals into the defined tasks allows for robots to manifest its thinking process when carrying out long-horizon plans. Moreover, these tasks are provided as natural language questions and answers, and can be viewed as a series of Visual Question Answering (VQA) steps. This formulation is similar to chain-of-thought for language model prompting~\cite{wei2023chainofthought}.  We also note concurrent work~\citep{hu2023thought} which demonstrates that mimicking step-by-step human thought improves 
planning accuracy.

\section{Models} \label{sec:result}

\subsection{\modelname}\label{subsec:methods}

We train a new model called \modelname{} derived from the \textbf{VideoCoCa} model ~\cite{yan2023videococa}, which is a video language model extending CoCa ~\cite{yu2022coca}. It uses an encoder-decoder architecture combining contrastive pretraining (like CLIP \citep{radford2021learning}) as well as generative pretraining (like SimVLM \citep{wang2022simvlm}) between video and text modalities. Unless otherwise stated, we use a VideoCoCa base model of 383M parameters with the initial checkpoint trained on image-captioning tasks as the original paper did, and fine-tune the model on the \algname{} video-text datasets. We choose a video-conditioned model to explore the importance of video in answering the visual questions in our dataset and find substantial benefits to video conditioning (see \fig{frames_error_reduction} and \ref{fig:error_rate_different_frames}).

\subsection{Baselines}\label{subsec:methods}

To compare with our finetuned model, we consider the following state-of-the-art baselines which have similar capabilities in visual question answering and planning for robotics.

\textbf{PaLM-E}~\cite{driess2023palme} is a visual language model built from pretrained ViT \cite{chen2023pali} and PaLM~\cite{chowdhery2022palm} LLM models, which projects images into the token embedding space of the pretrained LLM.  In our experiments we test PaLM-E-562B \textit{zero-shot}, without training on \algname{} dataset. While not finetuning is not a head to head comparison of models, the point of this comparison is establish how well state-of-the-art models trained on prior datasets can perform in the real world, and motivate further scalable data collection efforts to address the remaining performance gap.

\textit{Planning Methods.} We experiment with four baseline planning methods: two of which use \modelname{} and PaLM-E (zero-shot), as end-to-end planning models.  As two other baselines, we adapt the methods of \textbf{SayCan}~\cite{saycan2022arxiv} and \textbf{Grounded Decoding}~\cite{huang2023grounded}, which use a text-only LLM (PaLM~\cite{chowdhery2022palm}) in either phrase-level or token-level decoding guided by a visual affordance function (using \modelname{} as a video value function for affordance).

\begin{figure*}[h]
  \centering
  \includegraphics[width=.8\linewidth]{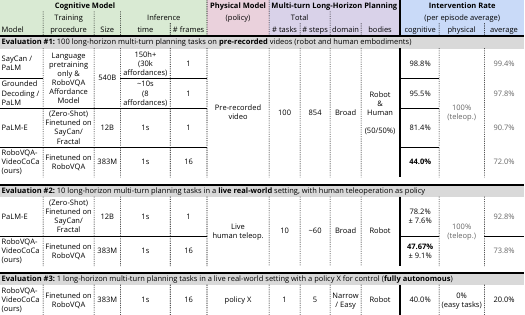}  
\caption{\textbf{Planning benchmarks with Intervention}: evaluation \#1 evaluates 854 planning steps on long-horizon episodes from \algname{} dataset, evaluation \#2 is performed live on a robot teleoperated by a human, while evaluation \#3 is controlled end-to-end by our model and a policy. Note that thanks to human intervention in the loop, all tasks are performed to completion even when the model makes mistakes.
}
  \label{fig:results-baselines}
\end{figure*}

\section{Benchmarks} \label{sec:roboplanbench}

\subsection{VQA Benchmark}

We first evaluate the model performance on individual tasks, where each task consists of a video segment and a question. The inference result is compared using exact match against prior human evaluation results stored in a central database as correct/incorrect for the video-question pair. The inference results for which no match is found are then collected for human raters to evaluate. During evaluation, a human rater is presented with the exact video segment and question as presented to the model. The rater is asked to either mark the model-generated answer as correct or incorrect, in which case the rater can propose a correct answer. All answers are added to the database, with the correctness of each answer marked accordingly.

We report the error rate for all models in \fig{vqa_eval_all} and find that there remains a substantial gap in performance for zero-shot state-of-the-art models compared to the finetuned model. While this is not too surprising, it is a valid question to ask when seeing good qualitative results by recent VLMs. Here we quantitatively prove that further scalable data collection efforts are required when deploying in the real world. In this graph we also make the case for video conditioning over image conditioning by presenting substantial gains with the former.

\subsection{Planning Benchmark with Intervention}

\begin{figure*}[h]
  \centering
  \includegraphics[width=.7\linewidth]{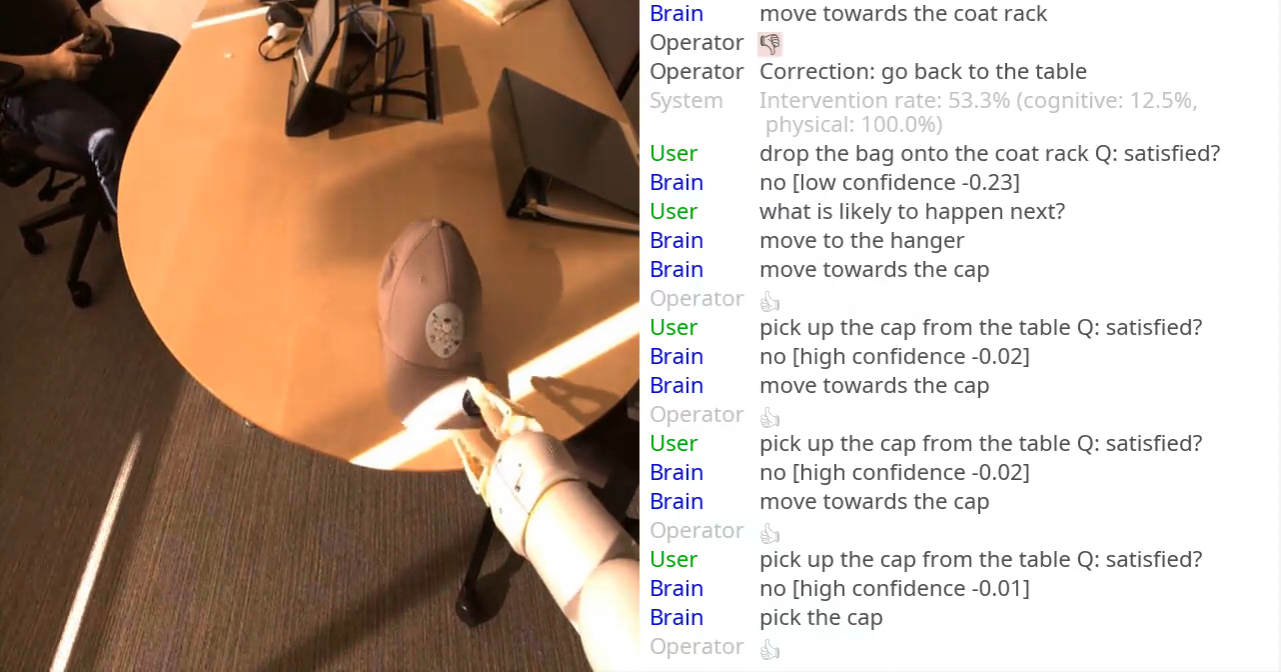}  
  \caption{\small{\textbf{Example of grounded chat with cognitive intervention}.
  Our model "Brain" is tasked with the following task at the beginning of the chat: "take the bag and cap on the desk and hang them on the coat rack" in this case. The bottom of the chat shows the most recent messages. The model is ran on an existing long-horizon video from the \algname{} dataset and produces medium-horizon plans to fulfill the long-horizon request. An operator is in the chatroom and validates each plan or provides a correction if incorrect. The user is also able to ask questions at any point in time. Here we see that the operator intervened and the system reported a cognitive intervention rate of 12.5\% at this point of the episode.
  }}
  \label{fig:chat}
\end{figure*}

\textbf{Intervention:} In \fig{results-baselines}, we propose 3 different evaluations of long-horizon planning. Each evaluation is measured by intervention rate, which we further decompose into \textit{cognitive} for the high-level text domain and \textit{physical} for the low-level motor command domain. However all progress can be measured with the single intervention rate which averages the cognitive and physical rates. This distinction is useful when physical actions are teleoperated (100\% physical intervention) to decouple high-level evaluations from low-level ones. Because the \algname{} dataset is very broad and diverse, we need an evaluation procedure that can test that entire breadth. Current low-level policies however tend to only perform in very narrow domains, this decoupling thus allows us to test the full breadth of tasks in evaluations \#1 and \#2. See \fig{chat} for an example of cognitive intervention in the chat window between the user, the model and the intervention operator.

\textbf{Offline Video Results:} In evaluation \#1, we run models on 100 long-horizon episodes (robot and human embodiments) from the \algname{} dataset which amounts to 854 planning steps in total. Models are given the long-horizon instruction and need to output medium-horizon plans, which are graded by humans. Note that the SayCan and Grounded Decoding baselines have slow inference time which makes them impractical to run in a live settings (hence not showing in other evaluations). Similarly, the inference time of the PaLM-E 562B model is too slow for real time (~30s), so we use a smaller version here. Note that despite being is 30x smaller, our model outperforms the state-of-the-art model by 46\%.

\textbf{Live Real-world Results:} In evaluation \#2, the high-level models are given a long-horizon instruction and provide medium-horizon plans in real time to a real robot teleoperated by a human. In evaluation \#3, a policy is deployed instead of a human teleoperator, but the domain is a lot narrower given the limited abilities of the policy. See videos of these evaluations at \website{}. While with evaluation \#3 we can obtain a much lower intervention rate thanks to the policy deployment, the domain is a lot narrower and emphasizes the need for a decoupled evaluation for high-level reasoning in broad domains.

\section{Analysis}

\subsection{Task Augmentation Matters}

\begin{figure}[h!]
  \centering
  \includegraphics[width=1\linewidth]{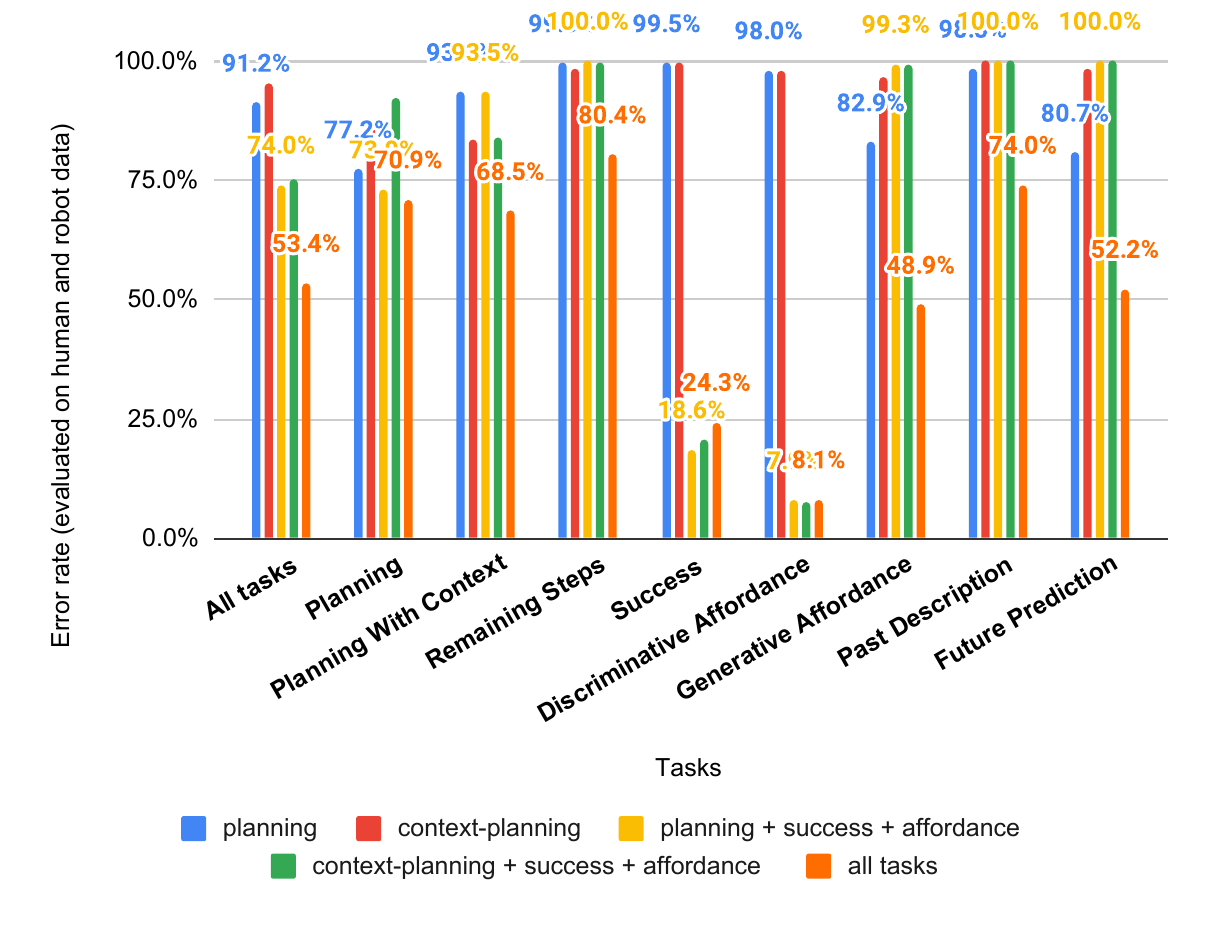}  
  \caption{\small{\textbf{Error rates for models trained with different sets of tasks}. Each model is trained and evaluated on the (robot + human) dataset, but using different subsets of tasks. We find that training on all tasks leads to better planning (70.9\% error) compared to training on planning only (77.2\% error).
  }}
  \label{fig:error_rate_different_tasks}
\end{figure}

In \fig{error_rate_different_tasks} we trained models on different following set of tasks: planning only, context-planning only, planning + success + affordance, context-planning + success + affordance, or all tasks. Note that when comparing planning vs. all tasks, the model trained on planning only sees 38M examples of planning task, while the one trained on all tasks sees roughly 1/8 the number of samples for the planning task. We find that the model trained on all tasks is often better of comparable than the models dedicated to a subset of tasks, with the exception of the success task. For example training on all tasks leads to better planning (70.9\% error) compared to training on planning only (77.2\% error). From a collection cost perspective, it is interesting to note that despite coming from the exact same set of instructions, the free tasks augmentation yields better results at no extra cost, hence task augmentation matters for performance and collection scalability.

\subsection{Tasks Transfer via Cross-Embodiment Data}

\begin{figure*}[h!]
  \centering
  \includegraphics[width=.9\linewidth]{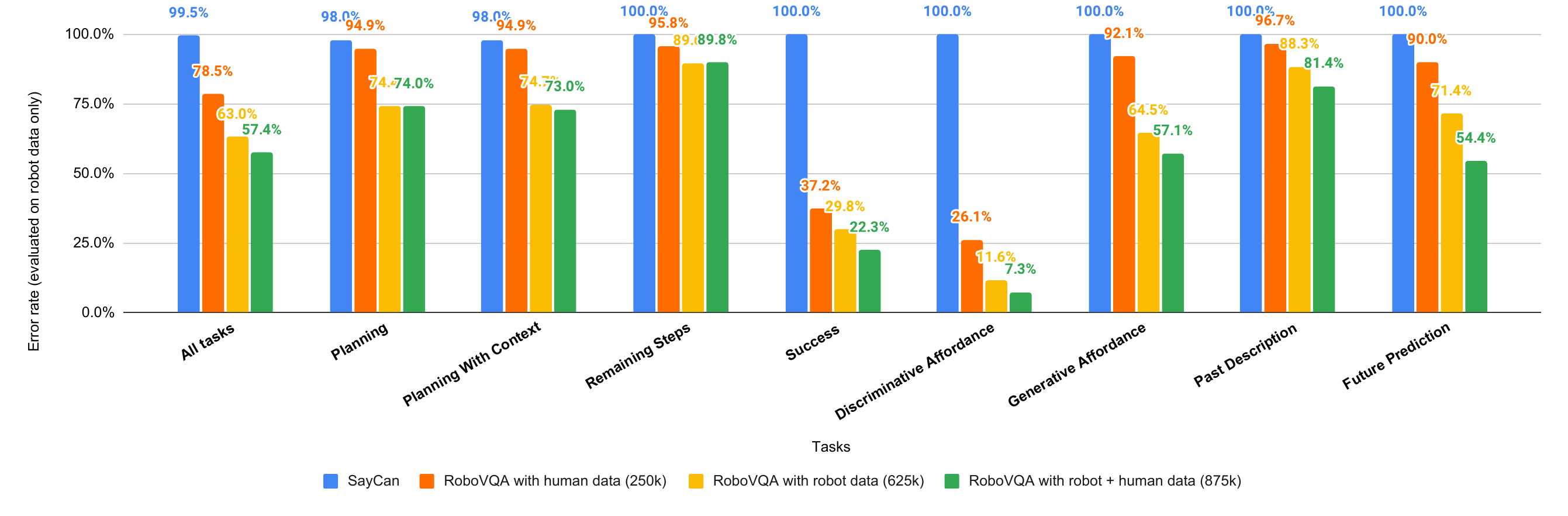}
  \caption{\small{\textbf{Error rates on robot-only test set}, comparing models trained on robot only, human only or both embodiments. We observed that while it is not trained on robot data, the model trained on human data still performs with less than 100\% error. We also find that the cross-embodiment training is beneficial even when evaluated on robot data only.
  }}
  \label{fig:vqa5_eval_robot}
\end{figure*}

\begin{figure*}[h!]
  \centering
  \includegraphics[width=1\linewidth]{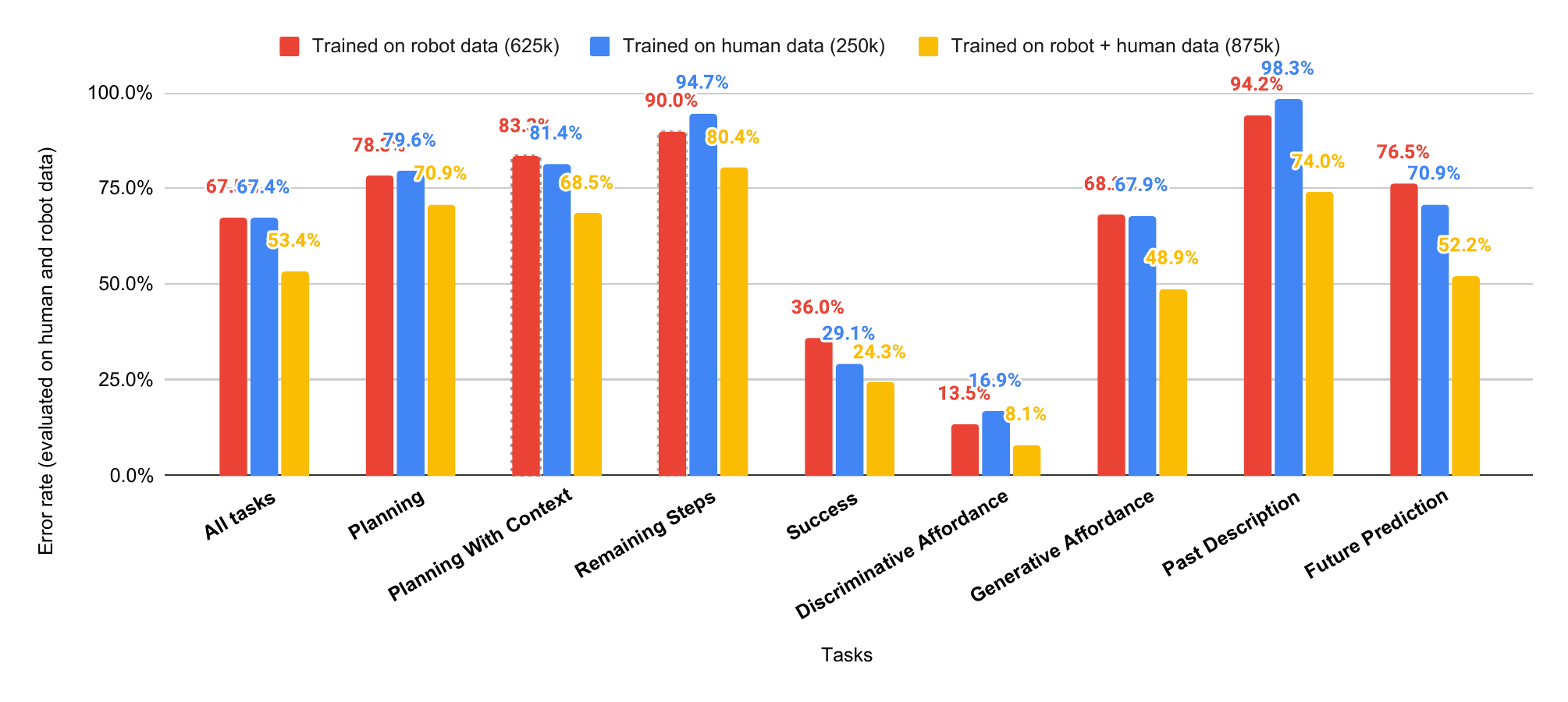}  
  \caption{\small{\textbf{Error rates on robot+human test set}. While it is expected that the model trained on both embodiments performs best, this graph illustrates that this model has the most breadth in capabilities and embodiments.
  }}
  \label{fig:vqa5_eval_robot_human}
\end{figure*}

In \fig{vqa5_eval_robot}, we compare error rates on the test split using \modelname{} trained on robot embodiment only, human embodiment only, and their combination. The test set contains only robot embodiment data. Despite cross-embodiment, we find that errors are below 100\% for all tasks when training on human data only, indicating human data by itself is useful to acquire a grounded understanding of videos with robot embodiment. Furthermore, training on both embodiments performs better than training on robot data only, indicating that extra data with human embodiment does not hurt performance when evaluating on the robot embodiment. We use~\citep{saycan2022arxiv} as a baseline, which uses a small, fixed list of 60 tasks and can only be evaluated on the planning task. We also provide the affordance answers from \algname{} as affordance function to SayCan for planning. Similarly, we evaluate on the joint human and robot test split in \fig{vqa5_eval_robot_human}. While it is not surprising that training on both embodiments performs best on the robot+human test set, we also shows it is the most general model as it performs better in all situations. More analysis is available in \sect{expemb}.

\subsection{Importance of Video modeling}

\begin{figure*}[!h]
  \centering
  \includegraphics[width=1\linewidth]{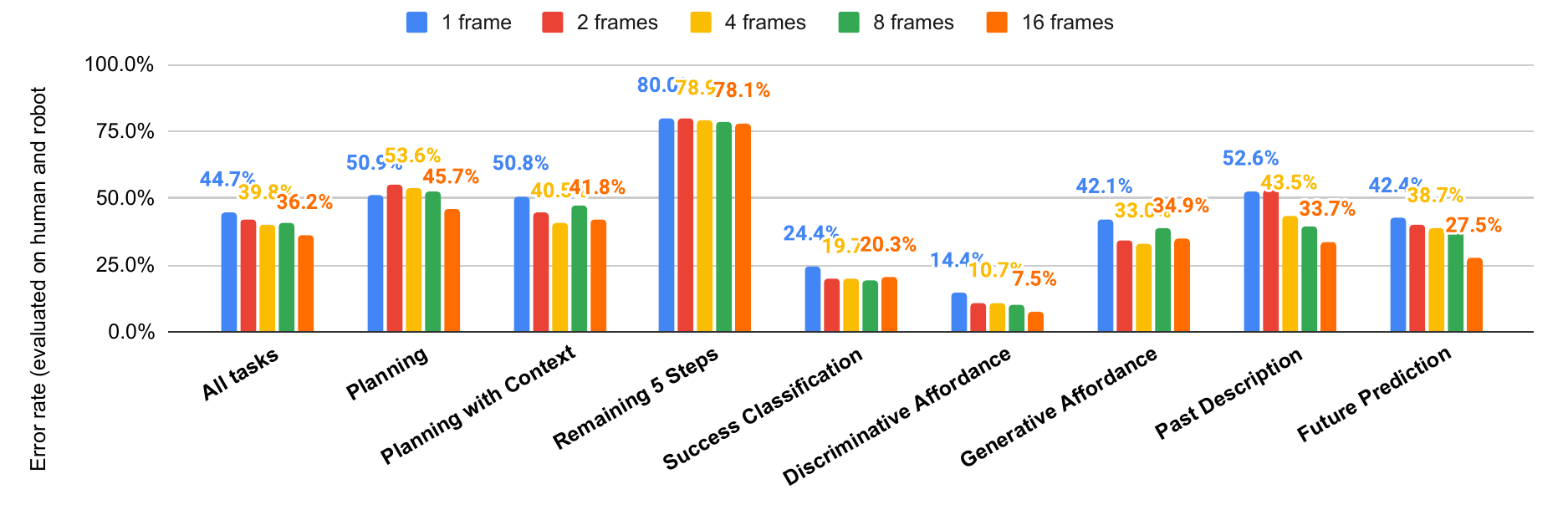}  
  \caption{\small{\textbf{Error rates for video model trained with different number of frames.} The model is trained on 875k samples (robot + human) and evaluated on the (robot + human) test set. We find that 16 frames yields the best results.
  }}
  \label{fig:error_rate_different_frames}
\end{figure*}

\begin{figure}[h]
  \centering
  \includegraphics[width=1\linewidth]{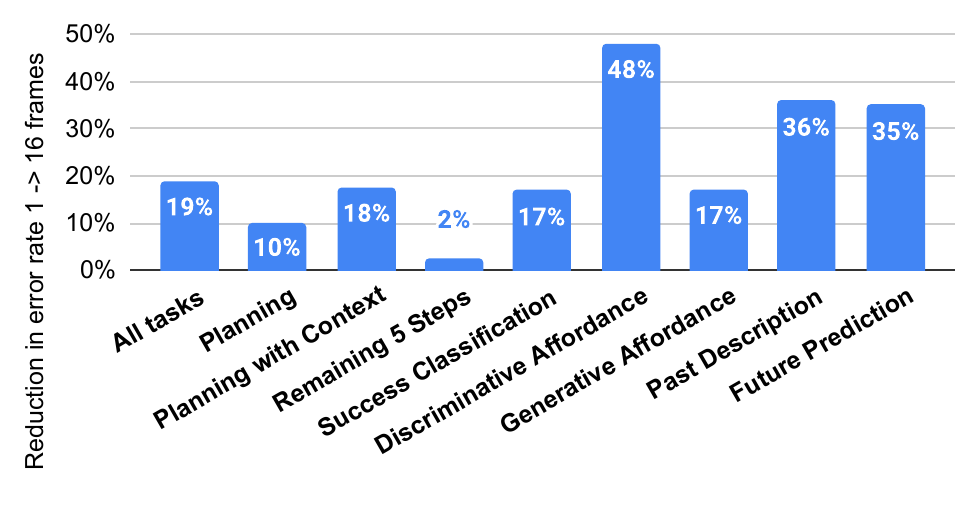}  
  \caption{\small{\textbf{Rate of error reductions when training a model with 16 frames as input versus 1  }}}
  \label{fig:frames_error_reduction}
\end{figure}

We investigate performance gains from video by training our model with (1, 2, 4, 8, 16) frames in \ref{fig:error_rate_different_frames} and find substantial error reductions in \fig{frames_error_reduction} between 1 and 16 frames. As expected, modeling with more frames yields better results, as it captures longer temporal dynamics for more accurate visual grounding.

\begin{figure}[h]
  \centering
  \includegraphics[width=1\linewidth]{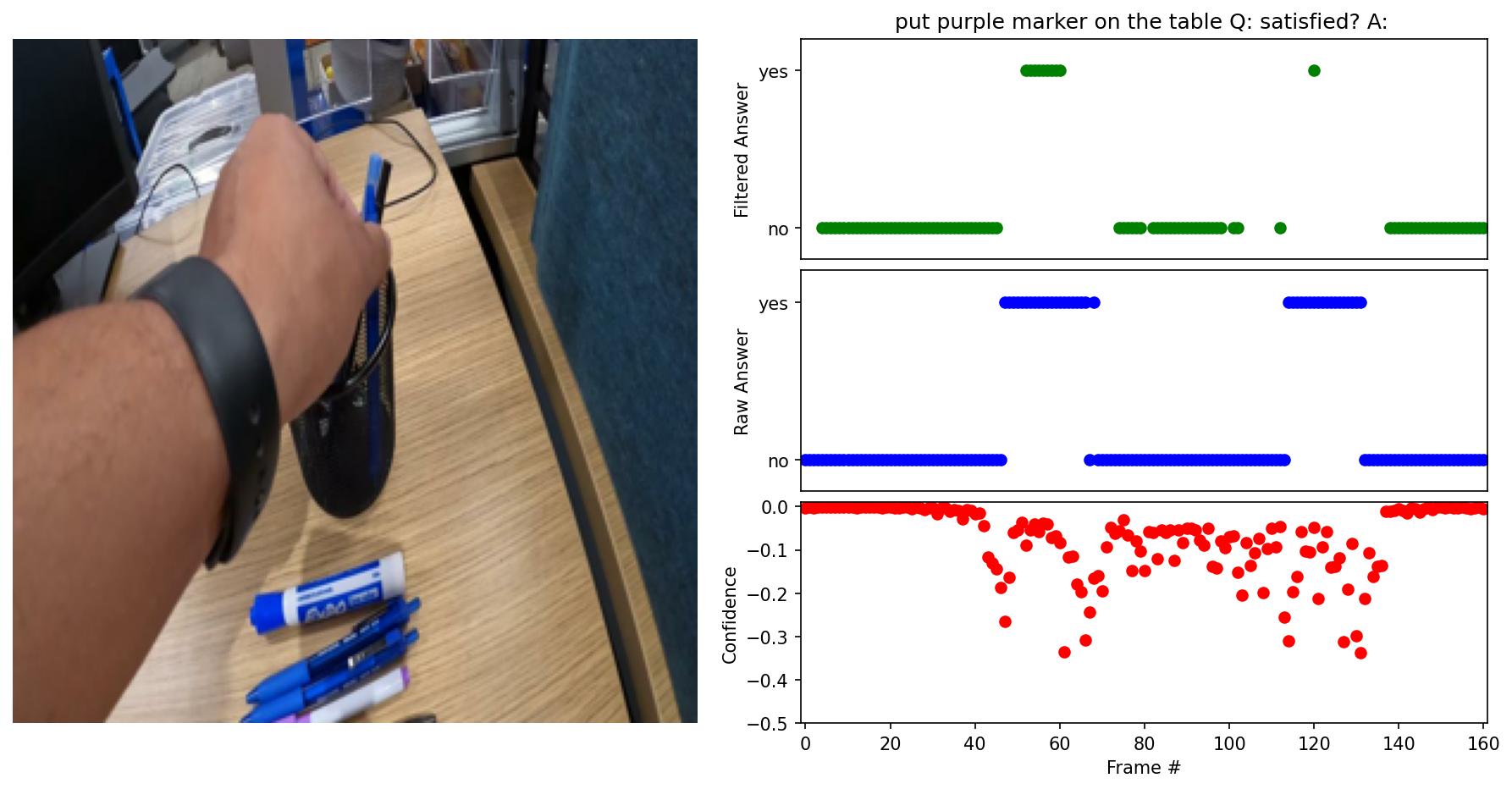}
  \caption{\small{\textbf{\modelname{} used for video success detection}. In blue are the raw answers to the question "put purple marker on the table Q: satisfied? A:", the confidence is shown in red and the answer filted by confidence is shown in green.}}
  \label{fig:success_detection}
\end{figure}

\subsection{Video Value-Functions}

We evaluate our model as a general grounded value-function from video and observe that it can provide stable binary detections as shown in \fig{success_detection}. Moreover, when filtering by the confidence of the yes/no tokens, we can further improve the accuracy of the success detection. These value functions can be used for closed-loop planning to know when a step is performed. Additionally, thanks to the dataset breadth and to video conditioning, the value functions can give richer understanding than traditional image-based success or affordance detectors.

\section{Related Work}

\noindent\textbf{Vision-Language Models.} Recently many methods~\citep{radford2021learning,jia2021scaling,li2022blip,yu2022coca,wang2022simvlm,gupta2022towards,chen2023pali} have been proposed that aim to train vision-language models (VLMs) on large-scale image-text pair datasets. We find the features learned by these methods generalize to robotic datasets. In this work, we also fine-tune a pre-trained vision language model called VideoCoCa~\citep{yan2023videococa} on conversation data grounded in long-horizon videos. The advantage of this VLM is that it is the encoder can consume full videos which helps in fine-grained temporal reasoning required to solve the tasks introduced in the \algname{} benchmark.

\noindent\textbf{Video Captioning.} Our task is closely related to the task of video captioning~\citep{wang2018video,gao2017video,pan2017video,luo2020univl, lin2022swinbert} which is a well studied problem in computer vision. In fact, we fine-tune a pre-trained video-captioning model VideoCoCa on these long-horizon videos. Different from the video captioning problem, all the videos in our fine-tuning dataset are egocentric. Also, we collect segment labels for a long-horizon task executed by either a robot or human. Furthermore, we augment these segments with a variety of question-answer pairs that add more supervision to the model so that an agent can execute long-horizon tasks. 

\noindent\textbf{Video Datasets with Text Annotations.} Recently many large-scale video datasets have been introduced~\citep{damen2018scaling,sigurdsson2018charades,lei2018tvqa,yu2019activityqa,miech19howto100m,yang2021justask,xiao2021next,grauman2022ego4d} that include videos of humans performing tasks with text narrations or question-answer annotations. Ego4D is the most similar dataset to the \algname{} dataset because Ego4D also has egocentric view of daily human activities annotated with dense narrations. However, our dataset differs in two key aspects. First, we collect human and robot interactions in the same environment. Second, our focus is on tasks that a robot is capable of doing. We hope that by lowering the domain gap between the human and robot videos we can achieve more transfer from human videos (which are faster to collect) to robot videos.
\cite{mees23hulc2} also explores scalable ways to collect language data with unstructured play~\cite{lynch2020language}, however they rely on an LLM requiring a prompt with a scene description that matches the environment’s state and is limited to 25 medium-horizon instructions.
Like \algname{}, TEACh\citep{teach} is another dataset that also contains interactive dialogues required to solve household tasks. However, TEACh consists of data in simulated environments while our dataset is collected in real kitchen and office environments with both humans and robots.

\noindent\textbf{Language Models for Planning.} \citep{huang2022lang} used a large language model (LLM) to produce plans for robotic tasks. This has been followed up by many works that also use LLMs to produce feasible next steps for a robot~\citep{saycan2022arxiv,driess2023palme,song2022llm,silver2022pddl,liu2023llm+}. One advantage of using LLMs to plan is that the output of these models can be used as input to language-conditioned policies~\citep{jang2021bc,rt12022arxiv,lynch2022interactive} that may have been trained independently.

\noindent\textbf{Intervention Rate.} Intervention Rate is a commonly used evaluation metric~\citep{steinfeld2006common,murphy2013survey,riedelbauch2023benchmarking} in robotics and self-driving car literature for measuring the performance of policies. In this work, we use it as a metric and as a mean to perform all tasks to completion, a necessary condition for real-world deployment.

\noindent\textbf{Chain of Thought Prompting.} \citep{ling-etal-2017-program,cobbe2021training,wei2023chainofthought} use the idea of prompting a language model with the process or steps to perform a reasoning task. The authors observe that prompting allows the model to improve performance on symbolic reasoning tasks like algebraic problems. Inspired by those results, we also provide rationale or thought supervision to the model by providing the sub-tasks as hindsight labels for successfully achieving the long-horizon task. 

\section{Limitations} \label{sec:limitation}

Some long-horizon episodes may be too repetitive and easy, thus we have filtered out episodes with more than 5 identical medium-horizon steps. Subsequently we observed gains in generalization.
Additionally we have not compared the effectiveness of the proposed human-and-robot dataset/benchmark with human-only dataset/benchmarks like Ego4D~\citep{grauman2022ego4d}, EpicKitchens~\citep{Damen2018EPICKITCHENS} etc., which merit careful study in our future work.

\section{Conclusion} \label{sec:conclusion}
We have shown a long-horizon collection approach with higher throughput and high diversity and breadth and released the resulting dataset for the benefit of the robotics community. We have demonstrated on real robots a number of capabilities learned with this dataset and established planning benchmarks with intervention as a metric and as a means for deployment.

\section*{ACKNOWLEDGMENT}
\small{
We thank Tran Pham, Elio Prado, Gavin Gonzalez, Jodilyn Peralta, Joseph Dabis, Alex Luong, Jodexty Therlonge, Rochelle Dela Cruz, Brianna Zitkovich, Emily Perez, Eric Tran, Huong T Tran, Dee M, Utsav Malla, Samuel Wan, Justice Carbajal and Scott Lehrer, Joel Magpantay, Sarah Nguyen, Steve Vega and Jonathan Vela for their contributions to data collection.
}

\bibliographystyle{IEEEtranN}
\newpage
\bibliography{root}

\clearpage

\section{Appendix}

\subsection{Random frames from training set}

\begin{figure}[!h]
  \centering
  \includegraphics[width=1\linewidth]{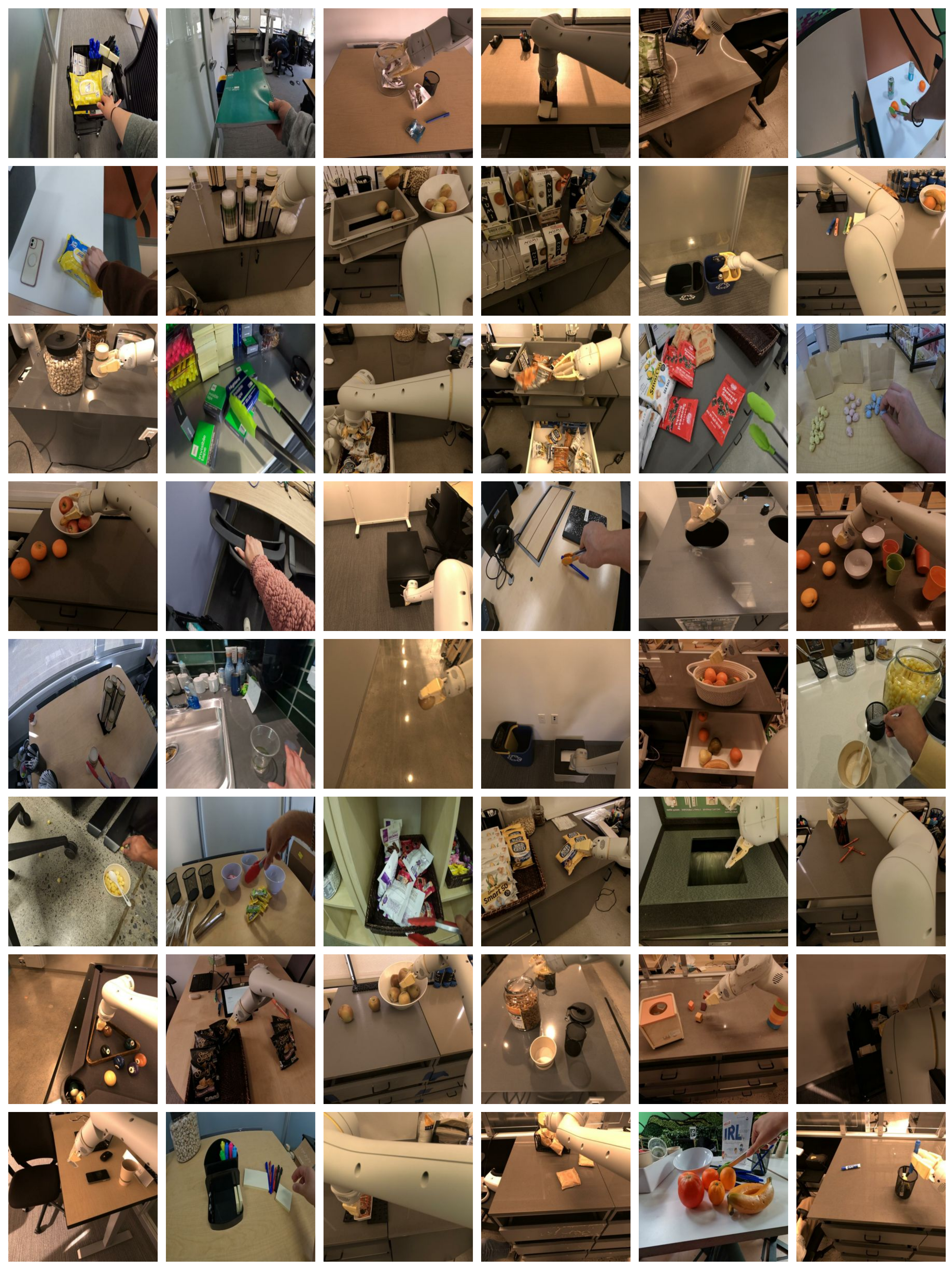}  
  \caption{\small{Random frames from training set.
  }}
  \label{fig:random_training_frames}
\end{figure}

\subsection{Dataset Statistics}
\label{sec:dataset_statistics}

\begin{figure}[h!]
  \centering
  \includegraphics[width=1\linewidth]{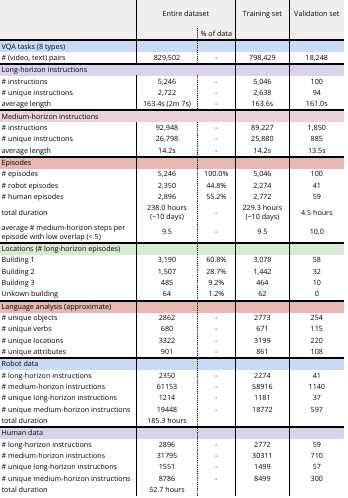}  
  \caption{\small{Dataset statistics.}}
  \label{fig:dataset_statistics}
\end{figure}

\begin{figure}[h!]
  \centering
  \includegraphics[width=1\linewidth]{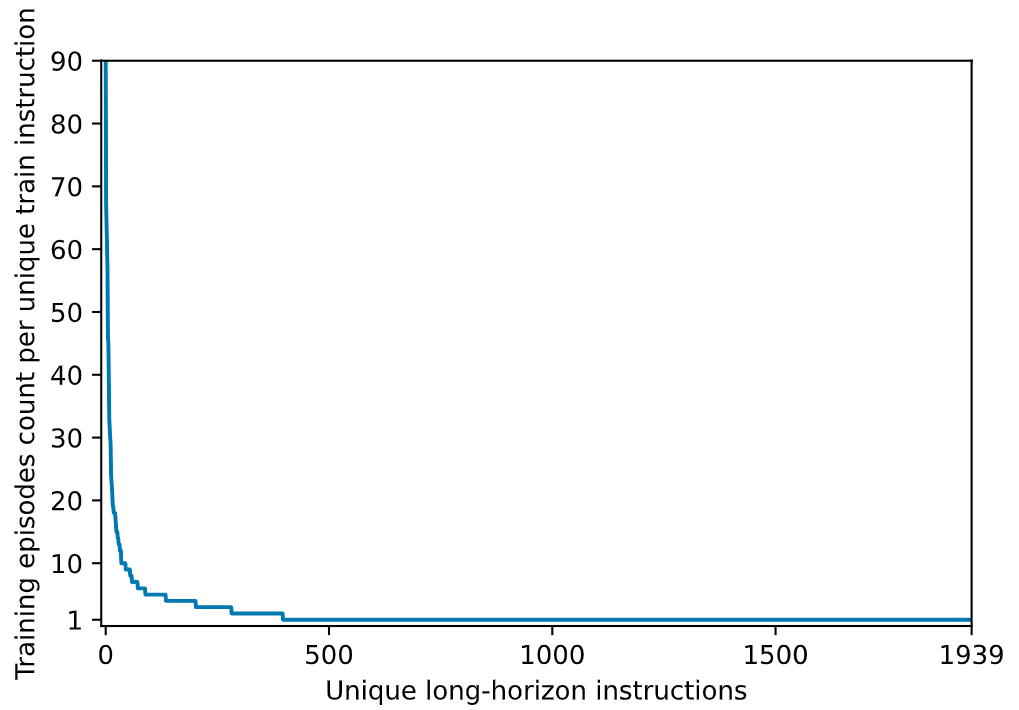}  
  \caption{\small{Number of training episodes per unique instruction: the maximum number of episodes for a unique long-horizon instruction is 90, the average 2.01 and the median is 1. There are 3894 training episodes which yield 1939 unique long-horizon instructions.
  }}
  \label{fig:train_episodes_per_train_instruction}
\end{figure}

\begin{figure}[h!]
  \centering
  \includegraphics[width=1\linewidth]{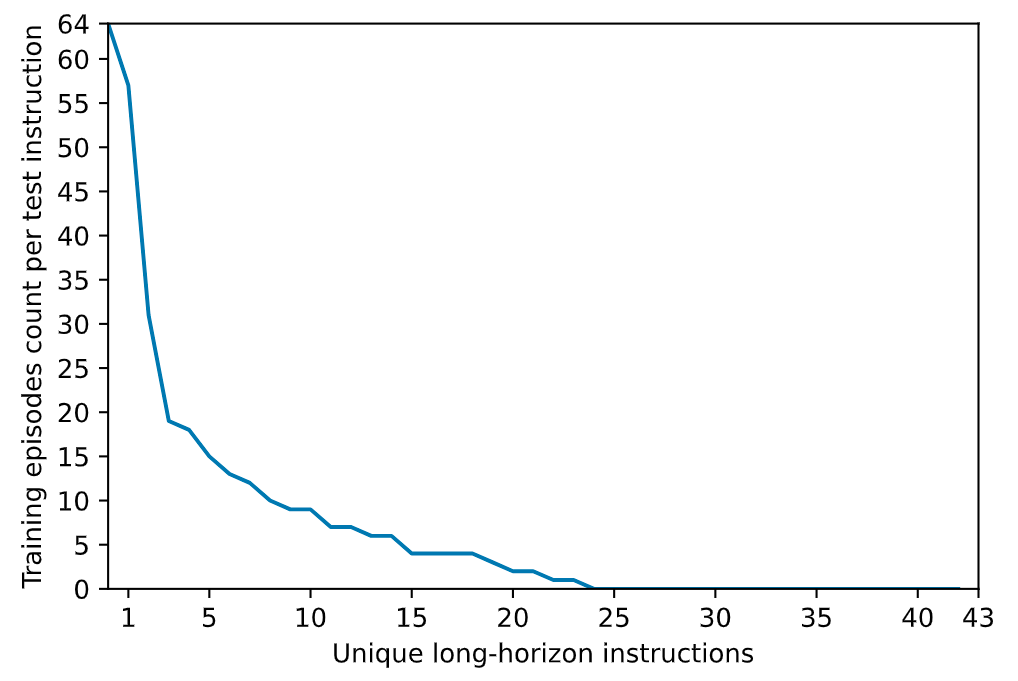}  
  \caption{\small{Number of training episodes that have the same long-horizon instruction as a test episode. Test episodes were sampled randomly and hence follow a similar distribution as observed in \fig{train_episodes_per_train_instruction}. Among the 43 episodes in the test set, we find that 23 of them have at least one episode with the same long-horizon instruction in the training set. For 20 of them (46\% of the test set), the long-horizon instruction is not present in the training set.
  }}
  \label{fig:train_episodes_per_val_instruction}
\end{figure}

As reported in \fig{dataset_statistics}, the entire dataset is a collection of 5246 long-horizon episodes (5046 for training and 100 for validation). Each episode has 1 long-horizon instruction and a varying number of medium horizon instructions that are temporally segmented. There are 2638 unique long-horizon instructions in the training set. Each unique long-horizon instruction has an average of 2.01 episodes collected, median is 1 and maximum is 90. See \fig{train_episodes_per_train_instruction} for the number of training episodes per long-horizon instruction. In \fig{train_episodes_per_val_instruction} we show the number of training episodes that have the same long-horizon instruction as a test episode. We find that 46\% of the test episodes do not have a long-horizon match in the training set. We show random frames from the training set in \fig{random_training_frames} and random long and short horizon instructions from the training set in \ref{appendix:instruction_samples}. We also provide extensive analysis of the language found in the training set in \ref{appendix:language_analysis} by automatically breaking down short-horizon instructions by categories (objects, actions, locations and attributes) using an LLM. This analysis found 2862 objects (e.g. ”tissue box”, ”purple color plate”), 680 skills or verbs (e.g. ”add something into something” or ”go out of a room”), 3322 locations or spatial relations (e.g. ”in the green plate”, ”left trash can”) and 901 attributes (e.g. shapes, color). Note that these numbers are only indicative as some objects can be redundantly described for example, see \ref{appendix:language_analysis} for more details.

\begin{figure*}[h!]
  \centering
  \includegraphics[width=0.8\linewidth]{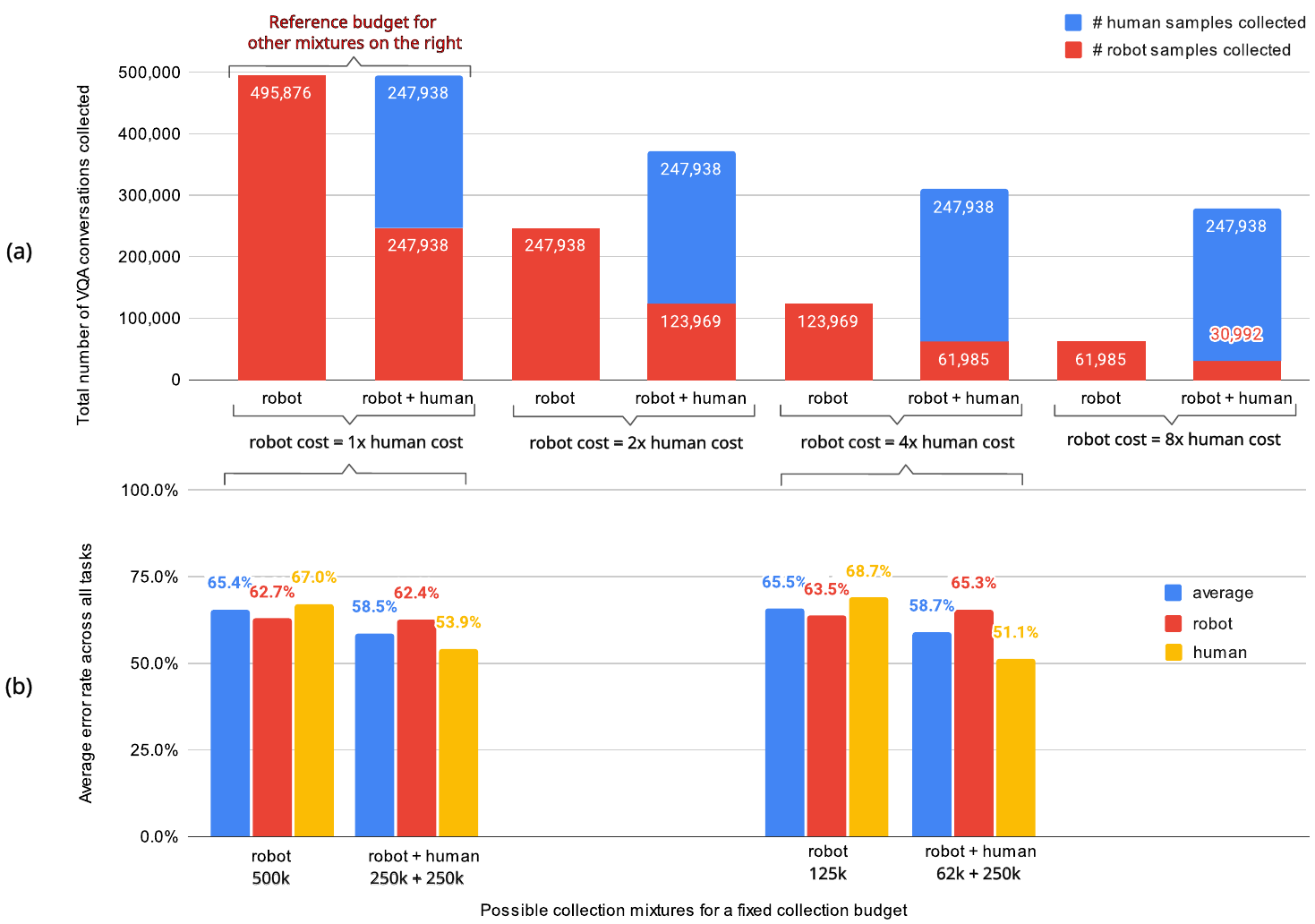}  
  \caption{\small{\textbf{Possible embodiment mixtures for a fixed collection budget.} This graph illustrates the possible trade-offs in total amounts of VQA samples collected for a fixed collecting budget and depending on the collection cost ratios between robot and human embodiments. In (a) we simulate different cost ratios by reducing the dataset size of the robot-embodiment dataset while keeping an equal budget for each embodiment. We calibrate this graph with a reference fixed budget that can produce approximately 500,000 VQA conversations at human collection cost. In (b) we report the error rates of each mixture (average error rate over all tasks). We find that mixing embodiments is overall beneficial even when the collection costs are the same and even when evaluating on the robot embodiment data only.
  }}
  \label{fig:collection_mixtures}
\end{figure*}

\subsection{Comparing Embodiment Mixtures} \label{sec:expemb}

Robot collection throughput will often be a factor of the cost including time, money, tele-operator training and availability, hardware maintenance etc., while humans are already expert of their own embodiment, collecting data with much less cost and cycle than robots. When factoring in all of these parameters into a collection budget, we can see that robot-to-human collection cost ratios and throughputs can vary wildly depending on all of these parameters. It is hence a critical question while scaling up data collection to know which data mixture for a given budget leads to the lowest error rates.

We explore this question in \fig{collection_mixtures} by looking at the data yields for a fixed collection budget of 500,000 VQA conversations, and report the performance for different configurations in Figure~\ref{fig:collection_mixtures}-b to analyze the trade-offs between different mixtures. We find that even if the robot-human ratio is 1.0 and only evaluating on the robot test set, the error rate is comparable when training on the equal robot250k-human250k mixture (62.4\%) compared to the full 500k robot dataset (62.7\%), while also being significantly lower on the human test set (53.9\% vs 67.0\%). Not only there is no downside for the robot performance to mix human data, it also makes the model more general and usable for other applications that require human embodiment understanding.

Similarly we find that when the robot-human cost ratio is 4.0, the performance of the mixed dataset (robot-62k + human-250k) on the robot test set is similar to the robot-only 125k dataset (65.3\% vs 63.5\%) while also being significantly lower on the human test set (51.1\% vs 68.7\%). We also observe that the performance gains seem rather small when training on 500k robot samples vs 125k, and that performance on human data degrades slightly when increasing robot data from 62k to 250k. We conclude that this analysis validates the common intuition that human data collection is an efficient way to scale up data collection for robots, despite the embodiment differences.

\subsection{Instructions Samples}
\label{appendix:instruction_samples}

We print 50 random instructions from the training set for both long-horizon and short-horizon below to get a sense of what the data looks like.

\textbf{50 long-horizon instructions:}
\begin{itemize}
\item please place all of the highlighters into the pen holder
\item please clean up the spill and put cup back on mouse pad
\item Please flip the bowls and pickup the yellow, pink and green candies from the floor and place them  in bowls. Then restock the chips into the bin.
\item please grab a small bin from the cart, place it on the table, put the red pens on the table in it, then put it back on the supply cart
\item empty the chips onto the counter
\item Please flip the bowls and pickup the yellow, pink and green candies from the floor and place them  in bowls. Then place the tongs into the bins.
\item Please flip the bowls and pickup the yellow, pink and green candies from the floor and place them  in bowls. Then pick up the tongs from floor and place in bins.
\item please clean up the pistachios spill on desk
\item I am feeling a little sick, can you please get me a covid test in the cabinet at the end of the building, as well as return it back onto my desk.
\item put fruit on the bookshelf
\item fill the bowl with apples
\item prepare a cup of coffee with the espresso machine.
\item place candies into middle bowl and blue chip bag in left bowl
\item place items from counter to bin
\item I don't want the water anymore. Can you pour the water into the sink and then throw the cup away
\item move items from table to cart
\item can you take the wireless mouse box out of the filing cabinet and put it on top of the table for me
\item I am done using the room can you turn off all the lamps.
\item Tidy up the mk table by straightening out the fruit labels, lining up the utensil holders and straightening the honey bottle platform
\item there is rubbish on the table, please throw them away into the correct places in the disposal bins on the floor by the door
\item i'm done writing in my notebook, please close it up and return the pen to the pen holder
\item please bring my green shopping bag from the coat rack to the table
\item separate the toys and microfiber cloths into different baskets.
\item please remove the chips from the bowl and place them in the top draw.
\item I am done drinking the coffee can you throw it in a trash can and get me some laffy taffy from MK kitchen to my desk.
\item please put the sugar packets in the tray
\item Can you refill my water cup and replace the cap and straw?
\item Restock the Numi tea boxes into the correct places
\item put the chips in the bin.
\item put all the snacks in the tray.
\item move the mouse box from the Whitney conference room to the dining booth
\item Please place the cookie squares into the tray.
\item please stock caddy for phone room 
\item pick the apple out of the jar and take it to phone room 2a3
\item place only the green pears in the bowl
\item Restock the ice packs and bandage rolls
\item put all the screwdrivers in the cup
\item please get the colored plastic cups from the top drawer and put them on the countertop
\item empty bin onto the table
\item open locker 17. then bring bag of chips from desk 2p2a to locker. close locker 17.
\item throw away the cocunut water
\item Put the red pens in the cup and bring them to a table in the mk, then bring the large postit notes to the table also
\item make a virtal line of the plants and sort them by hight
\item please pick up the trash on the table and throw it away into the compost
\item bring a usb c charger from the bookshelf to the desk in the whitney room
\item take out duck from plate on counter in a group
\item put duck into the basket
\item i'm finished with this hint water, please go recycle it in the micro kitchen for me and then bring me back a bag of lesser evil popcorn, cheese flavor
\item Please flips the bowls then seperate the green, yellow and pink candy. Then remove the tongs and the forks from bins and place them on table.
\item put the fruits in the basket
\end{itemize}

\textbf{50 medium-horizon instructions:}
\begin{itemize}
\item Touch the green bag
\item go away from the table
\item Grab the tissue
\item place the banana into the small bowl
\item drop the cups on the table
\item place strawberry hint water bottle in the tray
\item place green marker in the cup
\item Drop the green candy packet in the container
\item Place the black book on the table
\item Pick the bag on the table
\item Arrange the  white packet in tray
\item open the cap of jar
\item place the  yellow packet in glass
\item Put the tilted cup up right on the table
\item Release the orange marker into the left holder
\item Turn to the right
\item drop yellow candy into the left bowl
\item place the cup backward
\item drop the blue pen on a table
\item open the white box
\item Put orange bowl in the box
\item place tissue in the tray
\item Put the banana on the white table
\item move away from the rack
\item place 2 pistachio in the vessel
\item move away from the hanger
\item Place the square symbol in the baby pink box
\item Move your arm towards the right chair
\item place the lead on the glass
\item Put the paper bag in the black container
\item put paper clip in the rectangular stand
\item move to the orange packet
\item throw the tissue paper in dustbin
\item Place the red pen on the table
\item move towards the apple
\item Move away from the hint bottle
\item Go to the right side chair
\item Place the left indoor plant on the table
\item draw R on board
\item put sugar packets in the container
\item Place the 2 red packets on the table
\item move to the orange cable on the table
\item Drop the white pebble in the transparent glass
\item drop the black container in the box
\item Draw a diagonal line from left
\item place the black cart to the corner
\item Put blue cup on the table
\item drop the apple on the floor
\item Place the red can in fridge
\item pick the sanitizer
\end{itemize}

\newpage
\subsection{Dataset Language Statistics Analysis by LLM}
\label{appendix:language_analysis}

We use an LLM to extract different attributes from each short-horizon instruction from the training set and find:
\begin{itemize}
    \item 1795 objects, e.g. "tissue box", "purple color plate".
    \item 494 actions, e.g. "add something into something", "go out of a room". 
    \item 2064 locations, e.g. "in the green plate", "left trash can".
    \item 462 attributes, e.g. shapes, color.
\end{itemize}

Note that no clustering is performed and these lists contain redundant descriptions for each categories, the counts above are not meant to represent unique instances.
In subsequent sections we display the full lists for each category above along with their parent categories inferred by the LLM.

\addtolength{\textheight}{-12cm}   

\end{document}